\documentclass[letterpaper, 10 pt, conference]{ieeeconf}

\IEEEoverridecommandlockouts 

\usepackage{cite}
\usepackage{siunitx}
\usepackage{amsmath,amssymb,amsfonts}
\usepackage{algorithmic}
\usepackage{graphicx}
\usepackage{textcomp}
\usepackage{xcolor}

\usepackage{booktabs}
\usepackage{multirow}
\usepackage{caption} 
\newcommand{\eat}[1]{}

\title{\LARGE \bf
Bio-inspired Rhythmic Locomotion in a Six-Legged Robot
}

\author{Advait Lonkar, Sarthak Khoche, and Shrisha Rao
\thanks{International Institute of Information Technology Bangalore,
26/C Electronics City, Bangalore 560 100, India}%
}

\begin{document}

\maketitle
\thispagestyle{empty}
\pagestyle{empty}


\begin{abstract}

Developing a framework for the locomotion of a six-legged robot or a hexapod is a complex task that has extensive hardware and computational requirements. In this paper, we present a bio-inspired framework for the locomotion of a hexapod. Our locomotion model draws inspiration from the structure of a cockroach, with its fairly simple central nervous system, and results in our model being computationally inexpensive with simpler control mechanisms. We consider the limb morphology for a hexapod, the corresponding central pattern generators for its limbs, and the inter-limb coordination required to generate appropriate patterns in its limbs. We also designed two experiments to validate our locomotion model. Our first experiment models the predator-prey dynamics between a cockroach and its predator. Our second experiment makes use of a reinforcement learning-based algorithm, putting forward a realization of our locomotion model. These experiments suggest that this model will help realize practical hexapod robot designs.  

\end{abstract}



\section{Introduction}

Research on legged robots has gained much attention in recent years \cite{springer}. The evolution of such robots has progressed rapidly, with many companies, institutions and universities involved \cite{boston}. However, the construction of such robots still remains a challenging task. Past studies \cite{wall-climb} \cite{softrobot} suggest that the performance of such robots improves when their design imitates the biological principles observed during the natural locomotion of insects. In this paper, we present a bio-inspired framework for rhythmic locomotion of a six-legged robot, drawing inspiration from the fundamentals of an insect's locomotion. We base our locomotion model on that of a cockroach, which comprises of a central nervous system and six limbs with three degrees of freedom each. A cockroach's central nervous system consists of relatively a very small number of neurons compared to mammals \cite{Cohen1967}. However, it is still able to control the movement of its six limbs and is efficiently able to coordinate motion in them. As is evident in nature, cockroaches are very agile and are able to run very fast \cite{fast-cockroach}. 

Researchers have tried to gain insights from existing biological systems and have used these insights to construct bio-inspired mechanical designs, specifically robots \cite{lizard} \cite{Koh517}. Design of legged robots that can navigate in unseen terrains is heavily inspired by insects \cite{insect-inspired}, specifically cockroaches, stick insects and other small insects. Wu \emph{et al.} \cite{Wueaax1594} demonstrate the construction of insect-scale fast-moving robust robots that can carry loads, climb slopes and have the sturdiness of a cockroach. However they do not incorporate the central nervous system aspect for the locomotion model of a cockroach. Nelson \emph{et al} \cite{hexapod} present the design and simulation of a cockroach-like hexapod robot, which incorporates the \textit{limb morphology} of a cockroach by providing multiple degrees of freedom in each limb. These previous works in the field of insect-inspired neuro-robotics present a practical realization of a six legged robot or a hexapod. To make such a hexapod robot function, a complex design of hardware, computation and control mechanisms is needed. In this paper, we present a robotic framework with low computational power, that provides efficient control over its movements.

Our locomotion model draws inspiration from the biological principles that govern a cockroach's locomotion. The inclusion of central pattern generators help rhythmically generate patterns in the limb joints, which leads to the transition of limb phases. And incorporation of an inter-limb coordination algorithm help efficiently coordinates transitions between the limbs. Thus it gives us more control over the movement generation with less overhead. 

We present a bio-inspired robotics framework together with a mathematical construct of a six-legged robot's locomotion model (Section~\ref{locomotion_model}). First, the locomotion model is fundamentally based on the morphology of the cockroach limb which has three joints. Each joint can be in one of its two respective states. So a limb phase is represented as a combination of the states of its three joints.
Second, the \textit{central pattern generators} (or CPGs) are the neural networks that produce oscillating and rhythmic patterns in the robot limbs \cite{cpg}. CPGs play the role of the controller of switching the robot limb phases, which at any point of time can be one of the two---\textit{swing} or \textit{stance}. Hence the CPGs are a central feature of our locomotion model. 
Third, the CPGs in our locomotion model employs an algorithmic \textit{inter-limb coordination} to coordinate the pattern generation in the six robot limbs \cite{royalsociety}. As a result of this, we present a holistic robotic framework for the six-legged robot's locomotion model to give more control over its movement generation.

We designed two experiments to validate our locomotion model (Section~\ref{experiments}). For our first experiment, we created a \textit{spiking neural network} \cite{TAVANAEI201947} that closely mimics the biological aspect of the central nervous system of a cockroach that is responsible for generating an escape response, when in the proximity of a predator \cite{Titlow2013NeuralCR}. With this experiment, we model the predator-prey dynamics that occur between a cockroach and its predator (toad), based on experimental findings presented by Camhi \emph{et al.} \cite{camhi}. The spiking neural network that captures this predator-prey dynamics yields an escape response corresponding to the cockroach-toad pair, which is then passed along to the locomotion model. The motion attribute in the form of a direction vector is used to simulate the motion using our locomotion model(Section~\ref{res2}). These simulated motion from these attributes is used to validate the novelty of our work, which is in form of the presence of state transitions, pattern generations and limb coordination in our locomotion model (Section~\ref{res1})

Our second experiment establishes a practical realization of our locomotion model, projecting it as a robotics framework to carry out tasks like path traversing, object tracking etc. We made use of \textit{reinforcement learning} (RL) algorithms for simulating motion of cockroach in unseen environments \cite{sichkar-a}. The RL model yields coordinates of the path to be undertaken, which is then again passed along to the locomotion model, to validate the locomotion model.

By integrating RL models with our locomotion model, we put forward our locomotion model as a practical framework for a six legged robot.

\section{The Locomotion Model} \label{locomotion_model}

Locomotor patterns in the limbs of a hexapod are rhythmic and cyclic in nature \cite{cpg}.  Such locomotion consists of two phases: a \textit{swing} phase---in which the limb needs to propulse through its medium of travel; and a \textit{stance} phase---in which the limb needs to return to its starting position to generate the next \textit{swing} phase.  A complex rhythmic pattern is generated in the limb by neural networks within the central nervous system of such organisms, called \textit{central pattern generators}~\cite{cpg}. This pattern generation is also dependent on the sensory feedback reported through the actual external movements.  However, the pattern generation task becomes even more complex when several multi-segmented limbs need to be controlled.  Since walking and running in a hexapod is based on coordinated movement of its limbs, it requires \textit{inter-limb coordination} to generate appropriate patterns in the limbs \cite{royalsociety}.

In this section, we describe the way in which we implemented the locomotion model for a hexapod.  Our implementation involves the six limbs for a hexapod, their corresponding central pattern generators, and the inter-limb coordination required to generate appropriate patterns in those limbs.  Alongside the description, we provide a mathematical formulation as a proof of concept of our model.  The following subsections describe the details of each of the modules.

\subsection{The Six Limbs} \label{sec_limbs}

The limb morphology that we implement is similar to that of a cockroach.  Each limb has three main joints, which allow different types of movements: the \emph{thoraco-coxal} (TC) joint allows \textit{protraction} and \textit{retraction} of the limb, the \emph{coxa-trochanteral} (CTr) joint allows \textit{levitation} and \textit{depression} of the limb and the \emph{femur-tibia} (FTi) joint allows \textit{flexion} and \textit{extension} of the limb.\\
The six limbs for a hexapod are represented by:
\begin{equation*}
     C_L = \{LH, RH, LM, RM, LF, RF\}
\end{equation*}
The limb-state $L_i$, where $i$ is any particular iteration of the movement generation, is defined as the 3-tuple of the states of the 3 main joints of the limb. 
\begin{equation}
    L_i = (TC, CTr, FTi) \label{li}
\end{equation}
Each of the joints can be in two possible states, so the total combinations of limb-states for three main joints would be 8.  But the morphology of the limb only allows 4 of the possible states~\cite{cpg}, as shown below:
\begin{align*}
    s_1 &= [\mathrm{retraction}, \mathrm{levitation}, \mathrm{extension}],\\
    s_2 &= [\mathrm{protraction}, \mathrm{levitation}, \mathrm{extension}],\\
    s_3 &= [\mathrm{protraction}, \mathrm{depression}, \mathrm{extension}],~ \text{and}\\
    s_4 &= [\mathrm{retraction}, \mathrm{depression}, \mathrm{flexion}]
\end{align*}


This means that the limb-state is forced to be in one of the above 4 states, i.e.: 
\begin{equation}
    L_i \in \{s_1, s_2, s_3, s_4\} \label{state}
\end{equation}
Here, $s_1$ and $s_2$ belong to the \textit{swing} phase, and $s_3$ and $s_4$ belong to the \textit{stance} phase. 

\subsection{Central Pattern Generator} \label{sec_cpg}

The complex rhythmic motor pattern is generated by neural networks within the central nervous system, called \textit{central pattern generators} (CPG).  The sensory signals from the limb joints activate the CPGs to generate the appropriate state in the main joint of the associated limb.

We have implemented CPGs associated with each of the six limbs.  Any limb follows the sequence $s_1 \rightarrow s_2 \rightarrow s_3 \rightarrow s_4$ and repeats, for it to be able to make the hexapod walk forward.  The CPG is responsible for generating the next state for its associated limb.  So the limb-state at any iteration is just the next state in order of the limb-state in the previous iteration.
\begin{align*}
    L_i = \mathrm{next}(L_{i-1})
\end{align*}

However, all the limbs cannot be in the same state at the same time. Any multi-limbed organism relies on adaptive coordination of limbs during walking.  The next subsection describes the inter-limb coordination~\cite{royalsociety} implemented for the 6 limbs.

\subsection{Inter-Limb Coordination} \label{sec_ilc}
Limbs can be coordinated mechanically based on the transfer of body load from one limb to another. The unloading coincides with a switch from the stance to swing states.  States $s_1$ and $s_2$ belong to the swing phase of the limb movement, and states $s_3$ and $s_4$ to the stance phase.

Each of the six limbs of the hexapod are labeled as being a right (R) or left (L) and a front (F), middle (M) or hind (H) limb.  Limbs are coordinated in a back-to-front pattern.  Stance phases of the laterally adjacent limbs overlap to transfer the load efficiently.  In our implementation, the patterns in the limbs are generated as shown in~\eqref{ilc2}: the patterns generated are dependent on the previous state of the limb and the updated pattern is next in sequence.
\begin{align}
    \mathrm{ilc}(L_i) = \mathrm{next}(L_{i-1}) 
    \label{ilc2}
\end{align}

$\mathrm{next}(L_i)$ is defined as the next limb-state in \eqref{state}.  For example, $\mathrm{next}(s_1)$ is $s_2$, $\mathrm{next}(s_2)$ is $s_3$, $\ldots$, and $\mathrm{next}(s_4)$ is $s_1$ again.  Thus the function $\mathrm{next}$ is a circular iteration over~\eqref{state}. 

This implementation of the pattern generation along with the inter-limb coordination results in a sequence pattern for all six limbs (Section~\ref{res3}).

\section{Experiments} \label{experiments}

We designed two experiments to thoroughly test our locomotion model and its different aspects.  These experiments were designed in a plug-and-play manner, where the experimental setup would provide motion attributes like coordinates and direction of motion to the locomotion model, which it then uses to periodically generate and update the patterns and states of each limb (per Section~\ref{locomotion_model}). Since our locomotion model is inspired by the agility of a cockroach, our first experiment is designed in a manner that replicates the cockroach’s motion dynamics in the presence of a predator.  Our second experiment is an instance of path planning in an unseen environment using reinforcement learning, modeled to present a practical realization of our locomotion framework.

\subsection{In the Presence of a Predator} \label{exp1}

The first experiment mimics the neural activities that take place inside a cockroach when it is in the proximity of a predator. This neural model is based on theoretical and experimental findings presented by Titlow \emph{et al.}~\cite{Titlow2013NeuralCR} and Camhi \emph{et al.}~\cite{camhi}.

The reason for choosing the structure of a cockroach to base our hexapod model on is primarily because of its agility and quick evasive behavior.  The ability of a cockroach to quickly detect a predator in its surroundings and respond to the imminent threat posed by the predator is attributed to a reflex circuit that consists of the \textit{cerci} and giant fiber system.  The cerci (plural of cercus) are a pair of horn-like, wind-sensitive structures located at the end of the abdomen.  The giant fiber system comprises of giant inter-neurons, 5 pairs of \emph{abdominal ganglia}, a \emph{ventral nerve cord} (VNC), and \emph{thoracic ganglia}.  Of these, the thoracic ganglia are primarily responsible for activating the running procedure based on the escape response. 

In experiments conducted by Camhi~\textit{et al.}, it was observed that in response to the strike by a toad, a cockroach makes an initial pivot away from the origin of the strike, by making the initial movement of the \emph{metathoracic} limb (a cockroach’s back limb, that are primarily responsible for moving the cockroach forward) in the opposite direction to the toad’s strike.  As part of these experiments, the relationship between the angle of the strike (of the toad) and the pivot angle (of the cockroach) was documented. 

\begin{figure}[!hbtp]
    \centering
        \includegraphics[scale=0.25]{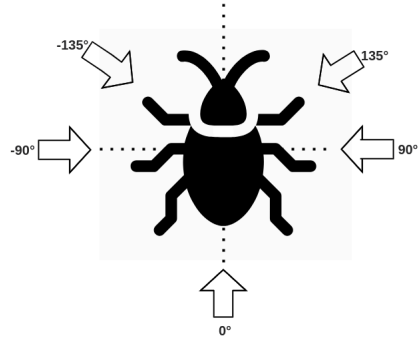}
        \caption{Angle convention}
    \label{fig:fig11}
\end{figure}

We model this predator-prey dynamics by incorporating this relationship between the angle of the strike and the pivot angle in our neural model.  The angle convention that we use in our model is given in Figure~\ref{fig:fig11}.  

The convention states that if the toad strikes from the \textit{southern} direction, then the angle of strike is to be considered as 0\si{\degree} or if the cockroach pivots towards \textit{eastern} direction, then the pivot angle is to be considered as +90\si{\degree}.

Based on this convention, we define \textit{stimulus\_angle} as the angle between a vector originating from the robot, and a vector originating from a predator, which essentially captures the angle of a predator's strike.  We define \textit{angle\_of\_turn} as the angle between the vector originating from a robot after the pivot and the initial vector originating from the robot, essentially quantifying the turn that a robot makes in order to pivot away from the predator. 

\begin{figure}[!hbtp]
    \centering
        \includegraphics[scale=0.38]{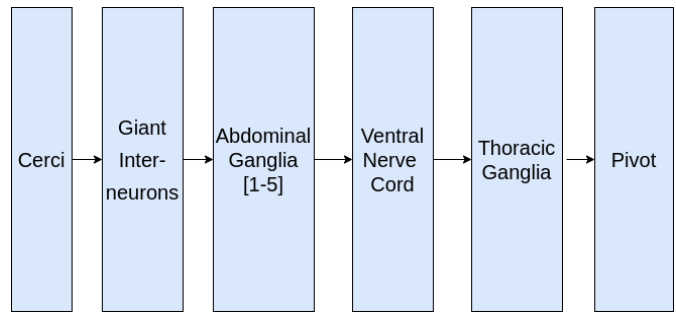}
        \caption{Model architecture}
    \label{fig:fig12}
\end{figure}

Now based on existing theoretical understanding~\cite{Titlow2013NeuralCR} of a cockroach's giant fiber system, we create a neural network architecture that is capable of modeling this dynamic between a cockroach and its predator.  To model this, we use \textit{spiking neural networks}~\cite{TAVANAEI201947} as these are biologically more similar to a biological nervous system and how it operates. The architecture of our spiking neural network is given in Figure~\ref{fig:fig12}.

The spiking neural network used to model the cockroach's reflex circuit is a simple feed-forward network, where each component in the nervous system is modeled as an ensemble of \textit{leaky integrate-and-fire} (LIF) neurons.  The input to this network is the \textit{stimulus\_angle}, which is captured by \textit{cerci}, a simple neural node that converts the mathematical value of \textit{stimulus\_angle} to its corresponding neural representation, which is then passed along the network. The network predicts the \textit{angle\_of\_turn} of the robot, which is captured by the \textit{pivot} neural node and converted back to its corresponding mathematical value for human interpretation. The weights are initialized randomly, and the loss used while training is the Mean Absolute Error (MAE) loss.

\subsection{In the Absence of a Predator} \label{exp2}

Our second experiment is an instance of path planning by a six-legged robot in an unseen environment, with the help of reinforcement learning (RL)~\cite{sichkar-a}.  In any RL setup, there are two entities---an agent and an environment.  At each step, the agent executes some action, based on which it receives some reward from the environment, in the form of a new state.  Each action by the agent is thus associated with a reward, and the goal of the agent is to take actions such that it maximizes the cumulative reward. 

The objective function of our RL model is as follows: 

\begin{center}
    $Q_{s', a'} = Q_{s, a} + \alpha [R(s, a) + \gamma*maxQ_{s, a} - Q_{s, a}] $
\end{center}

where \textit{s} is the current position of the agent and \textit{a} is current action taken by the agent. $Q_{s, a}$ is the value of objective function at the current state and $Q_{s', a'}$ is the value of objective function at the next state. $\alpha$ is the learning rate and $\gamma$ is the discount rate. R(s, a) is reward at current state and $maxQ_{s, a}$ is the maximum expected future reward. 

\begin{figure}[!hbtp]
    \centering
        \includegraphics[scale=0.40]{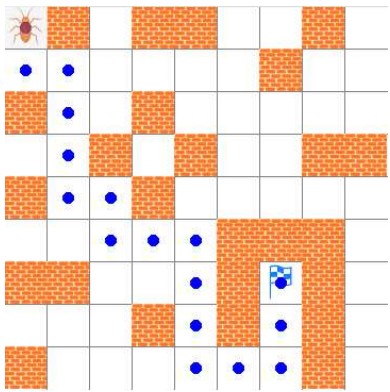}
        \caption{Agent, environment and Final trajectory}
    \label{fig:fig13}
\end{figure}

In our designed experiment, the robot is the agent that acts on the environment, with an objective to reach the destination from the source, while maximizing the rewards at each step.  The task is maze navigation, as can be seen in Figure~\ref{fig:fig13}.

The 9$\times$9 maze in Figure~\ref{fig:fig13} represents our environment, with walls as obstacles. The goal of the agent (robot) is to reach from the source (top-left corner) to the target (marked by the flag), while avoiding obstacles (walls) and maximizing the cumulative reward.  Our reward scheme is as follows: 

\begin{align*}
    \mathrm{Reward} = 
    \begin{cases}
      +1 & \text{if the agent moves closer to the} \\
         & \text{destination} \\    
      -1 & \text{if the agent hits an obstacle} \\
      0 & \text{if the agent sits idle}
    \end{cases}
\end{align*}

Based on the reward scheme and the objective function, the final Q-table is as shown in Table~\ref{t1}.  The final path taken by the agent corresponds to the maximum cumulative reward at each step.  The next direction corresponding to maximum reward is highlighted in bold in Table~\ref{t1}.  The final path undertaken by the agent corresponds to all the directions is highlighted in bold.

\begin{table}[!hbtp]
    \centering
    \resizebox{9cm}{!}{
    \begin{tabular}{lllll}
        \textbf{Coordinates} & \textbf{Up} & \textbf{Down} & \textbf{Right} & \textbf{Left}\\
        \toprule
        (0.0, 0.0) & 3.100243e-14 & \textbf{1.677161e-11} & -7.894016e-01 & 4.961888e-14 \\
        (0.0, 40.0 & 8.970498e-14 & -6.883389e-01 & \textbf{1.378912e-10} & 9.595142e-13 \\
        (40.0, 40.0) & -4.417339e-01 & \textbf{1.168131e-09} & 1.046329e-13 & 3.214153e-13 \\
        (40.0, 80.0) & 4.481458e-12 & \textbf{9.441269e-09} & 1.234715e-12 & -3.573884e-01 \\
        (40.0, 120.0) & 2.626601e-11 & \textbf{7.898185e-08} & -2.221786e-01 & 4.292473e-13 \\
        (40.0, 160.0) & 1.702751e-09 & 1.849776e-11 & \textbf{5.157568e-07} & -1.312542e-01 \\
        (80.0, 160.0) & \textbf{1.570568e-01} & 3.671630e-06 & -7.725531e-02 & 5.305350e-11 \\
        (80.0, 200.0) & 4.333286e-10 & 1.011155e-09 & \textbf{2.410241e-05} & 5.771410e-12 \\
        (120.0, 200.0) & -1.136151e-01 & 9.070283e-07 & \textbf{1.407129e-04} & 2.963661e-07 \\
        (160.0, 200.0) & 1.684900e-08 & \textbf{8.020513e-04} & -1.224790e-01 & 2.697013e-10 \\
        (160.0, 240.0) & 1.343483e-08 & \textbf{4.082327e-03} & -6.793465e-02 & 3.032200e-07 \\
        (160.0, 280.0) & 2.883518e-05 & \textbf{1.678626e-02} & -4.900995e-02 & -4.900995e-02 \\
        (160.0, 320.0) & 2.452179e-05 & 1.122709e-03 & \textbf{6.081326e-02} & 3.483398e-06 \\
        (200.0, 320.0) & -8.648275e-02 & 2.749313e-03 & \textbf{1.767578e-01} & 7.224155e-04 \\
        (240.0, 320.0) & \textbf{4.175657e-01} & 4.253015e-03 & -2.970100e-02 & 3.535026e-03 \\
        (240.0, 280.0) & \textbf{7.829551e-01} & 3.604843e-03 & -1.990000e-02 & -6.793465e-02 \\
        \bottomrule
    \end{tabular}}
    \caption{Final Q-table}
    \label{t1}
\end{table}

\begin{figure}[!hbtp]
    \centering
        \includegraphics[scale=0.50]{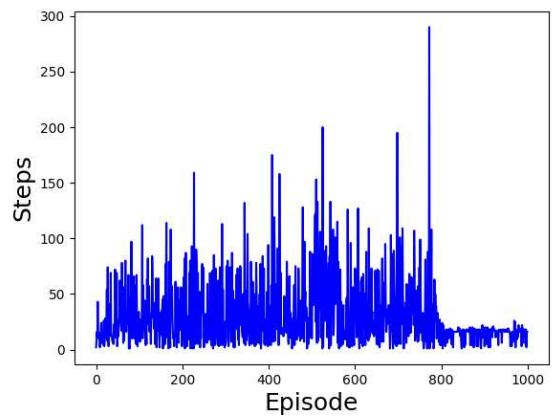}
        \caption{Number of episodes vs Number of steps}
    \label{fig:fig14}
\end{figure}

\begin{figure}[!hbtp]
    \centering
        \includegraphics[scale=0.50]{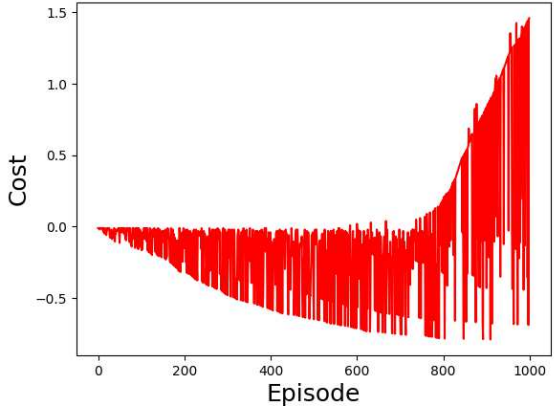}
        \caption{Number of episodes vs Cost of each episode}
    \label{fig:fig15}
\end{figure}

The graphs corresponding to the agent's interaction with the environment are given in Figures~\ref{fig:fig14} and~\ref{fig:fig15}. In these figures, an episode is a complete instance of the Q-learning algorithm from the start state to a terminal state.    

The final path undertaken by the agent based on the interaction with the environment, while maximizing the reward (dictated by the Q-table) is shown in blue dots in Figure~\ref{fig:fig13}.

\section{Results} \label{results}

In this section, we describe the results of our hexapod locomotion model design and the subsequent experiments carried out.  The section is divided into two parts---the first describes the hexapod's locomotion model's resultant features as described through the mathematical formulation in Section~\ref{locomotion_model}, and the second part describes the results of the experiments described in Section~\ref{experiments}.

\subsection{Hexapod Locomotion Model Design} \label{res1}
    \subsubsection{Hexapod limb's joints and states}
\begin{figure}[!hbtp]
            \centering
                \includegraphics[scale=0.25]{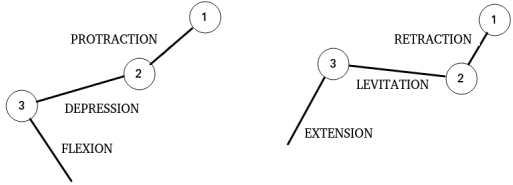}
                \caption{Hexapod limb's joints and states.}
            \label{fig:fig6}
        \end{figure}

        Shown in Figure~\ref{fig:fig6} is the hexapod limb design implemented for our locomotion model. The design of the limb stays consistent with the \eqref{li} given in Section~\ref{sec_limbs}. The joints are labeled as : 1 - TC joint, 2 - CTr joint, 3 - FTi joint.
        
In the first half of Figure~\ref{fig:fig6}, the state of the limb connection between joints 1 and 2 is $\mathrm{protraction}$, between joints 2 and 3 is $\mathrm{depression}$ and between the joint 3 and the free end of the limb is $\mathrm{flexion}$.  While the corresponding states in the second half of Figure~\ref{fig:fig6} are $\mathrm{retraction}$, $\mathrm{levitation}$, and $\mathrm{extension}$.  This is to highlight all the possible states of the individual limb connections.
    
\subsubsection{Patterns Generated in an Individual Hexapod Limb} 
        
\begin{figure}[!hbtp]
        \centering
            \includegraphics[scale=0.25]{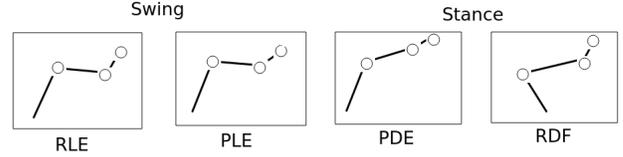}
            \caption{Patterns generated in an individual hexapod limb}
        \label{fig:fig7}
        \end{figure}

It can be inferred from Figure~\ref{fig:fig6} that each joint connection can be in two possible states, which leads to a total of 8 possible combination of states for the whole limb.  However, it so happens that there are only 4 possible combination of states allowed for a hexapod limb for it to move, as described in Section~\ref{sec_limbs}.  As shown in Figure~\ref{fig:fig7}, these are the 4 possible states that a hexapod limb can be in at any point of time.  The \textit{central pattern generator} described in Section~\ref{sec_cpg} is the module responsible to generate the rhythmic patterns as shown in Figure~\ref{fig:fig7}. 
        
In Figure~\ref{fig:fig7}, the hexapod limb's joints are denoted as RLE ($\mathrm{retraction, levitation, extension}$), PLE ($\mathrm{protraction, levitation, extension}$), PDE ($\mathrm{protraction, depression, extension}$) and RDF ($\mathrm{retraction, depression, flexion}$).  The hexapod limb is in the states denoted in Figure~\ref{fig:fig7} consistent with the \eqref{state} given in Section~\ref{sec_cpg}.
    
\subsubsection{Centrally coordinated patterns generated in the hexapod limbs} \label{res3}
        
The first two states of Figure~\ref{fig:fig7} belong to the \textit{swing} phase of the hexapod limb, while the next two states belong to the \textit{stance} phase.  The \textit{swing} and \textit{stance} phases of a limb are coordinated through the \textit{inter-limb coordination} described in Section~\ref{sec_ilc}. 
        
\begin{table}[h]
    \centering
    \begin{tabular}{lllll}
        \textbf{Limbs} & \multicolumn{4}{c}{\textbf{Limb phase}}\\
        \toprule
        LH & swing & stance & swing & stance \\
        LM & stance & swing & stance & swing \\
        LF & swing & stance & swing & stance \\
        \midrule
        RH & stance & swing & stance & swing \\
        RM & swing & stance & swing & stance \\
        RF & stance & swing & stance & swing \\
        \bottomrule
    \end{tabular}
    \caption{Coordinated pattern generation in all the hexapod limbs.}
    \label{t2}
\end{table}
    
The coordinated patterns generated in the hexapod limbs are as shown in Table~\ref{t2}, which is a snapshot of any particular 4 iterations of the state changes in all six limbs of the hexapod.  The limbs are labeled in the Table~\ref{t2} as \textit{LH, LM, LF, RH, RM, RF} where \textit{L} and \textit{R} stand for \emph{left} and \emph{right}, while \textit{H, M} and \textit{F} stand for \emph{hind}, \emph{middle}, and \emph{front} as described in Section~\ref{sec_ilc}.  The limbs clearly transition from \textit{swing} to \textit{stance} phases alternatively, generated by \eqref{ilc2} given in Section~\ref{sec_ilc}.

\subsection{Simulated Trajectory of Hexapod} \label{res2}

Both our experiments mentioned in Section~\ref{experiments} were designed in a plug-and-play manner, to directly fit and integrate with our locomotion model.  The purpose of our experiments is to test the locomotion model, by providing motion attributes like coordinates and direction of motion to the locomotion model.  Based on these motion attributes, the locomotion model then periodically updates states and patterns in each of the hexapod's limbs. 

The purpose of the first experiment was to model a cockroach's reflex response in the presence of a predator, to enable a suitable, similar set of actions in a hexapod robot.  This is done when the neural model is coupled with the inter-limb coordination algorithm of the locomotion model of the robot, and central pattern generators that emulate the cockroach's reflex circuit. 

The output of this experiment is in the form of \textit{pivot\_angle} which essentially quantifies the turn hexapod would make in order to get away from a predator. For the purpose of this experiment, we assumed some size-appropriate values for velocity and acceleration of the hexapod, while assuming the toad to remain stationary throughout the course of the experiment. To test our locomotion model, we created an instance of a Robot-Predator pair and simulated the predator-prey dynamics between them (as mentioned in Section~\ref{experiments}).  We simulated the motion of the hexapod as dictated by the predator's location and traced out its trajectory. While executing its motion, the hexapod also periodically updates the states of each limb in accordance with the inter-limb coordination mechanism (Section~\ref{sec_ilc}).   

In order to generate the trajectory, we had to \textit{iteratively} run the experiment mentioned in Section~\ref{exp1}, as it more closely resembles the behavior of a cockroach in an identical situation---a cockroach constantly re-assesses a predator's location, and based on that moves away from it, and keeps on doing this until it reaches a safe distance~\cite{camhi}. 

One iteration of our experiment (Section~\ref{exp1}) thus constitutes computation of \textit{stimulus\_angle} based upon the robot's and predator's locations. This \textit{stimulus\_angle} is passed as an input to our spiking neural network which gives us the \textit{angle\_of\_turn}. This \textit{angle\_of\_turn} dictates the direction the hexapod would move in.  The hexapod traverses in that direction with the assumed motion attributes for 1 time step, while internally generating patterns and changing states in each limb.  After each successive time step, it recomputes the \textit{stimulus\_angle} based on its new location and repeats the same procedure for several such iterations.  
 
\begin{figure}[!hbtp]
    \centering
        \includegraphics[scale=0.40]{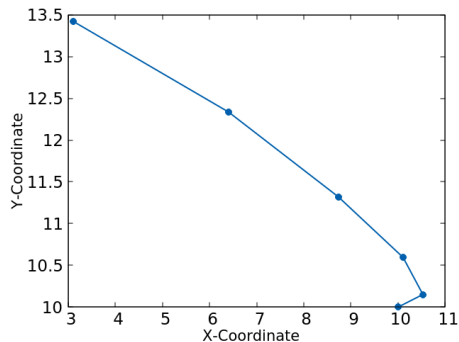}
        \caption{Simulated trajectory of the hexapod}
    \label{fig:fig16}
\end{figure}

The trajectory of the hexapod object in presence of a predator, traced out for five iterations, is shown in \textit{Fig. \ref{fig:fig16}}, which closely resembles the trajectory of an actual cockroach in presence of a toad~\cite{camhi}.

The purpose of the second experiment (Section~\ref{exp2}) was to present a practical realization of our locomotion model, where the coordinates of the trajectory are provided by a reinforcement learning (RL) algorithm.  While the motion attributes are pre-decided by the RL algorithm (Figure~\ref{fig:fig13}, the state changes that occur internally remain the same as dictated by the locomotion model (Figure~\ref{fig:fig7}, Table~\ref{t2}).

(Apart from RL, one could even use deep-learning or other techniques to dictate the motion of a hexapod, while using our locomotion model as the underlying framework for the movement of each of its limbs.)

\section{Conclusion}

We present a bio-inspired framework for the cockroach-like locomotion of a hexapod.  This is a modular framework which includes the central pattern generators and inter-limb coordination to generate appropriate patterns in the hexapod limbs.  Our experiments indicate that our framework, when coupled with appropriate hardware and computational units, can be used for robots that can carry out tasks like path traversal and object tracking.  As our framework uses simple bio-inspired control mechanisms, construction of such robots can be done in an efficient way, requiring a smaller amount of computational power. 







\bibliographystyle{IEEEtran} 
\bibliography{refs}

\end{document}